\documentclass{article}

\usepackage{arxiv}

\usepackage[utf8]{inputenc} 
\usepackage[T1]{fontenc}    
\usepackage{hyperref}       
\usepackage{url}            
\usepackage{booktabs}       
\usepackage{amsfonts}       
\usepackage{nicefrac}       
\usepackage{microtype}      
\usepackage{lipsum}
\usepackage{graphicx}
\graphicspath{ {./images/} }

\usepackage[utf8]{inputenc} 
\usepackage[T1]{fontenc}    
\usepackage{hyperref}       
\usepackage{url}            
\usepackage{booktabs}       
\usepackage{amsfonts}       
\usepackage{nicefrac}       
\usepackage{microtype}      
\usepackage{xcolor}         

\usepackage{subfigure}

\usepackage{graphicx}
\usepackage{mathptmx}      
\usepackage[T1]{fontenc}
\usepackage[utf8]{inputenc}
\usepackage{amsmath,amsfonts,amsthm,bm}
\usepackage{amssymb}
\usepackage{systeme}
\usepackage{longtable,tabularx}
\usepackage {graphicx}
\usepackage{caption}
\usepackage{floatrow}   
\usepackage{graphicx,xcolor} 
\usepackage[framemethod=tikz]{mdframed}
\usepackage{caption}

\usepackage{algorithm}%
\usepackage{algorithmic}%

\floatname{algorithm}{Algorithm}

\usepackage{xcolor}
\def\mathbi#1{\textbf{\em #1}}
\usepackage{caption}
\usepackage{sidecap}
\usepackage{breqn}
\usepackage{bm}
\usepackage[square,numbers]{natbib}

\usepackage{floatrow}
\def\mathbi#1{\textbf{\em #1}}

\newcommand{\DesignVar}{\mathbi{x}}

\newcommand{\DesignSpace}{\chi}
\newcommand{\ObjFun}{f}
\newcommand{\MinObjFun}{f^{*}}
\newcommand{\MinDesignVar}{x^{*}}
\newcommand{\LevFid}{l}
\newcommand{\MaxLevFid}{L}
\newcommand{\Dim}{d}
\newcommand{\IterOpt}{i}
\newcommand{\NoisyObserv}{y}
\newcommand{\IndNumObs}{n} 
\newcommand{\NumObs}{N} 

\newcommand{\StandDevNoise}{\sigma_{\epsilon}}
\newcommand{\NorDist}{\mathcal{N}}
\newcommand{\MeanFunGP}{\mu}
\newcommand{\CovFunGP}{\kappa}

\newcommand{\StandDevGP}{\sigma}
\newcommand{\KernelMatrix}{\mathbi{K}} 

\newcommand{\Kronecker}{\delta}
\newcommand{\ConsFact}{\varrho}
\newcommand{\Discrep}{\delta}
\newcommand{\AF}{U}
\newcommand{\EI}{EI}
\newcommand{\MFEI}{MFEI}
\newcommand{\Improv}{\gamma}
\newcommand{\CompCost}{\lambda}

\newcommand{\CumDistFun}{\Phi}

\newcommand{\Expectation}{\mathbb{E}}
\newcommand{\Stage}{z}

\newcommand{\Dataset}{S}
\newcommand{\ExpectedReward}{J}
\newcommand{\StageReward}{r}
\newcommand{\State}{s}
\newcommand{\Control}{c}
\newcommand{\Disturbances}{d}
\newcommand{\SysDyn}{\mathcal{F}}
\newcommand{\Policy}{\pi}

\newcommand{\NorVar}{Z}
\newcommand{\CholDec}{C}
\newcommand{\AuxMat}{H}

\newcommand{\MCiter}{j}
\newcommand{\Budget}{B}

\newcommand{\Error}{\Delta \ObjFun}

\title{Non-Myopic Multifidelity Bayesian Optimization}

\author{
 Francesco Di Fiore \\
  Politecnico di Torino\\
  \texttt{francesco.difiore@polito.it} \\
   \And
 Laura Mainini \\
  Massachusetts Institute of Technology\\
  Politecnico di Torino\\
  \texttt{laura.mainini@polito.it} \\
}

\begin{document}
\maketitle


\begin{abstract}
Bayesian optimization is a popular framework for the optimization of black box functions. Multifidelity methods allows to accelerate Bayesian optimization by exploiting low-fidelity representations of expensive objective functions. Popular multifidelity Bayesian strategies rely on sampling policies that account for the immediate reward obtained evaluating the objective function at a specific input, precluding  greater informative gains that might be obtained looking ahead more steps. This paper proposes a non-myopic multifidelity Bayesian framework to grasp the long-term reward from future steps of the optimization. Our computational strategy comes with a two-step lookahead multifidelity acquisition function that maximizes the cumulative reward obtained measuring the improvement in the solution over two steps ahead. We demonstrate that the proposed algorithm outperforms a standard multifidelity Bayesian framework on popular benchmark optimization problems. 
\end{abstract}

\let\thefootnote\relax\footnotetext{Full paper: Di Fiore, Francesco, and Laura Mainini. "NM2-BO: Non-Myopic Multifidelity Bayesian Optimization." Knowledge-Based Systems 299 (2024): 111959. https://doi.org/10.1016/j.knosys.2024.111959}


\section{Introduction}

Solving an optimization problem is a central task in disciplines ranging from engineering to operational research, to computer science and economics. Optimization problems are commonly formalized as the minimization of an objective function by iteratively choosing an input from a feasible set of variables. In many applications the function to optimize does not have a closed-form expression and its evaluation can be costly.

Bayesian optimization (BO) is a widely used strategy to solve optimization problems involving black-box objective functions that are expensive to compute \cite{MockusAl1978, SnoekAl2012}. BO uses a probabilistic surrogate model of the objective function to compute an acquisition function that is repeatedly maximized to select a new input closer to the optimum. Then, the objective function is observed at this new training sample, and used to update the surrogate model improving the accuracy of the prediction. BO requires evaluating the objective function several times to asses the solution of the optimization problem. This may result in a prohibitively computational expense for real-world applications requiring the use of lab-scale experiments or time consuming computer-based models, and where a limited number of evaluations of the objective are allowed to compute the optimization. 

In practice, many applications allow to use alternative representations at different levels of fidelity to model the objective function, where the lower fidelity levels tend to be less time consuming but produces less accurate results. Multifidelity methods target the efficient combination of multiple sources of information, using low-fidelity analysis to systematically reduce the number of high-fidelity evaluations through a continuous trade-off between cost and accuracy \cite{Kennedy&OHagan2000, ForresterAl2007, PeherstorferAl2018, BeranAl2020}. Multifidelity Bayesian optimization (MFBO) combines the function outputs at different levels of fidelity into a single probabilistic model, and use this information to compute a multifidelity acquisition function that allows to jointly select the new input location and the associated level of fidelity to query \cite{HuangAl2006, KandasamyAl2016, ZhangAl2017, TakenoAl2020}. Most widely-used MFBO algorithms are based on greedy approaches, where the multifidelity acquisition function measures only the immediate reward of the input selected at each step of the optimization \cite{FernandezAl2016, ParkAl2017, PoloczekAl2017, WuAl2020, LiAl2020}. However, the quantification of the long-term reward obtained at the future steps may lead to greater improvements in the search of the optimum, enhancing the computational efficiency of the predictive strategy. 

This motivates the interest for non-myopic acquisition functions that are capable to grasp the informative gains obtainable by looking ahead into future steps. In most existing non-myopic methods, the location of the new input is determined through a partially observable Markov decision process (POMDP) \cite{Ginsbourger&LeRiche2010, LamAl2016, Lam&Willcox2017}. However, the solution of the POMDP is computationally intense, limiting this approach to small-scale optimization problems \cite{Powell2007}. Several works proposed approximations of the POMDP to formulate two-step lookahead acquisition functions capable to enhance the performance of BO, while containing the computational expense that would arise looking more steps ahead \cite{OsborneAl2009, Ginsbourger&LeRiche2010, GonzalezAl2016, LamAl2016, Lam&Willcox2017}. However, these non-myopic formulations are conceived for a single fidelity Bayesian framework, which can be prohibitive for objective functions characterized by high computational demand. 

To address this issue, we propose a strategy for non-myopic multifidelity Bayesian optimization to maximize the long-term reward as an improvement of the optimal solution. Our framework is based on an original scheme for the maximization of the multifidelity acquisition function based on a two-step lookahead strategy, that allows to predict the informative gains obtainable at future steps of the optimization. We will refer to that as the two-step lookahead multifidelity acquisition function. Specifically, our contributions are:
\begin{enumerate}
    \item Formalize MFBO as a Dynamic Programming (DP) problem and define the best optimization policy that maximizes the cumulative reward obtained evaluating a specific input with a certain level of fidelity two steps ahead, quantifying its merit considering the long-term benefits.
    \item Formulate the two-step lookahead multifidelity acquisition function as the solution of the DP problem encompassing intractable nested expectations and maximizations problems whose closed-form solution is not available. To overcome this issue, the Monte Carlo approach is used to provide a robust estimate of the two-step lookahead multifidelity acquisition function. We also provide details about its integration in a multifidelity Bayesian settings to draw our non-myopic MFBO framework. 
    \item Illustrate and discuss the proposed Non-myopic MFBO in comparison with standard MFBO over benchmark problems specifically conceived to stress-test multifidelity methods. 
\end{enumerate}

The manuscript is organized as follows. Section~\ref{s: Problem statement and background} provides an overview of Bayesian optimization with Gaussian processes and its extension in a multifidelity scenario. Section~\ref{s: Non-myopic multifidelity active learning} presents our two-step lookahead multifidelity acquisition function and its implementation in a multifidelity framework. Then, in Section~\ref{s: Numerical experiments} the non-myopic MFBO is numerically investigated and compared to standard MFBO over a variety of benchmarks. Finally, Section~\ref{s: Conclusion} provides concluding remarks.

\section{Problem Formulation and Background}
\label{s: Problem statement and background}

In this paper, we consider a sequential decision problem in which we seek to minimize a black-box objective function $\ObjFun : \DesignSpace \rightarrow \mathbb{R}$ from noisy observations at iteratively selected sample points $\DesignVar \in \DesignSpace$ in the input space $\DesignSpace$. More formally, we can define our optimization problem as follows: 
\begin{equation} \label{e:OptPrb}
    \min_{\DesignVar \in \DesignSpace} \ObjFun (\DesignVar)
\end{equation}
In real world applications, the evaluation of the objective function is frequently expensive, therefore the interest in computational strategies to minimize the number of explicit calls of the function

\subsection{Bayesian Optimization via Gaussian Processes}

Bayesian optimization (BO) is a popular approach to solve the global optimization problem in Equation~\eqref{e:OptPrb} \cite{MockusAl1978, SnoekAl2012}. BO learns a probabilistic surrogate model to predict the values of the objective function across the input space and quantifies the uncertainty of the prediction. This information is used to compute an acquisition function that defines a policy to measure the improvement of the solution at different input locations, which usually encodes a trade-off between the exploration of the input space and the exploitation toward the optimum. BO selects the new input to query by maximizing the acquisition function, and updates the surrogate model with the observation of the objective function at the new point.

Commonly, BO relies on a Gaussian Process (GP) as the surrogate model to approximate the objective function \cite{Rasmussen2003}. The GP distribution $GP \left( \MeanFunGP (\DesignVar), \CovFunGP (\DesignVar, \DesignVar')  \right)$ is a joint Gaussian distribution over any finite set of $\NumObs$ random input points $\{ \DesignVar_\IndNumObs \}^{\NumObs}_{\IndNumObs=1}$ completely specified by its mean $\MeanFunGP(\DesignVar) = \Expectation \left[ \ObjFun(\DesignVar) \right]$ and covariance (also referred as kernel) function $\CovFunGP(\DesignVar, \DesignVar') = \Expectation \left[ (\ObjFun(\DesignVar) - \MeanFunGP(\DesignVar)) (\ObjFun(\DesignVar') - \MeanFunGP(\DesignVar')) \right]$. We assume that $\ObjFun \sim GP \left( 0, \CovFunGP (\DesignVar, \DesignVar')  \right)$ is the prior belief about the objective and that we have collected observations in the form $\{\DesignVar_\IndNumObs, \NoisyObserv_\IndNumObs \}^{\NumObs}_{\IndNumObs=1}$, where $\NoisyObserv_\IndNumObs \sim \NorDist (\ObjFun (\DesignVar_\IndNumObs), \StandDevNoise(\DesignVar_\IndNumObs))$ and $\StandDevNoise$ is the standard deviation of the normally distributed noise. 

The acquisition function $\AF : \DesignSpace \rightarrow \mathbb{R}^{+}$ under the GP prior depends on the previous observations of the objective and determines which point in $\DesignSpace$ should be evaluated in the next optimization step via maximization $\DesignVar_{new} \in \max_{\DesignVar \in \DesignSpace} \AF (\DesignVar)$. There is a variety of acquisition functions commonly adopted in literature, such as the expected improvement (EI) \cite{JonesAl1998}, upper confidence bound (UCB) \cite{SrinivasAl2009}, entropy search (ES) \cite{Hennig&Schuler2012}, max-value entropy search \cite{Wang&Jegelka2017} and predictive entropy search (PES) \cite{HernandezAl2014}. In this work we focus on GP based Bayesian optimization with the EI acquisition function, given its simple implementation and reasonable performances in practice: 
\begin{equation} \label{e:EI}
\begin{split}
     \EI(\DesignVar) =  
     \StandDevGP(\DesignVar)(\Improv(\DesignVar)\CumDistFun(\Improv(\DesignVar))) + \NorDist(\Improv(\DesignVar); 0,1))
\end{split}
\end{equation}
\noindent where $\Improv(\DesignVar) = (\ObjFun(\DesignVar^*) - \MeanFunGP(\DesignVar))/\StandDevGP(\DesignVar)$, $\DesignVar^*$ is the current location of the best value of the objective and $\CumDistFun(\cdot)$ is the cumulative distribution function of a standard normal distribution.

\subsection{Multifidelity Bayesian Optimization}
\label{s: Multifidelity Bayesian optimization}

Many applications allow multifidelity representations of the objective function $\{\ObjFun^{(1)},...,\ObjFun^{(\MaxLevFid})\}$, where the higher the level of fidelity $\LevFid$, the more accurate and costly the evaluation of $\ObjFun^{(\LevFid)} (\DesignVar)$. We indicate with $\MaxLevFid$ the highest-fidelity level, that is considered as the ground truth source of information about the objective function. Multifidelity Bayesian optimization learns a single probabilistic model combining low-fidelity and high-fidelity observations, and choose the location of the next input and the level of fidelity to query through a multifidelity acquisition function. The probabilistic surrogate model is defined extending the Gaussian process to a multifidelity scenario through an autoregressive scheme \cite{Kennedy&OHagan2000}: 
\begin{equation} \label{e:AutoRegScheme}
\ObjFun^{(\LevFid)} = \ConsFact \ObjFun^{(\LevFid - 1)} \left( \mathbi{\DesignVar} \right) + \Discrep^{(\LevFid)} \left( \mathbi{\DesignVar} \right) \quad \LevFid = 2,...,\MaxLevFid
\end{equation} 
\noindent where $\Discrep^{(\LevFid)} \sim GP(0, \CovFunGP^{(\LevFid)} \left( \mathbi{\DesignVar}, \mathbi{\DesignVar}'\right))$ models the discrepancy between two adjoining levels of fidelity, and $\ConsFact$ is a constant scaling factor. Hence, we can obtain the posterior mean $\MeanFunGP^{(\LevFid)} (\mathbi{\DesignVar})$ and variance $\StandDevGP^{2(\LevFid)} (\mathbi{\DesignVar})$ of the objective through the covariance matrix $\KernelMatrix(i,j) = \CovFunGP \left( \left( \mathbi{\DesignVar}_{i}, \LevFid_i \right), \left( \mathbi{\DesignVar}_{j}, \LevFid_j \right) \right) + \Kronecker_{ij}\StandDevNoise(\DesignVar_i)$ and $ \CovFunGP_\NumObs^{(\LevFid)} (\mathbi{\DesignVar}) = \CovFunGP \left( \left( \mathbi{\DesignVar}_{i}, \LevFid_i \right), \left( \mathbi{\DesignVar}, \LevFid \right) \right)$:
\begin{gather}
    \MeanFunGP^{(\LevFid)} (\mathbi{\DesignVar}) = \CovFunGP_\NumObs^{(\LevFid)} (\mathbi{\DesignVar})^T \KernelMatrix ^{-1} \mathbi{\NoisyObserv} \label{e:MFGPmean} \\[10pt]
    \StandDevGP^{2(\LevFid)} (\mathbi{\DesignVar}) = \CovFunGP \left( \left(\mathbi{\DesignVar}, \LevFid\right), \left( \mathbi{\DesignVar}, \LevFid \right) \right) - \CovFunGP_\NumObs^{(\LevFid)} (\mathbi{\DesignVar})^T \KernelMatrix^{-1} \CovFunGP_\NumObs^{(\LevFid)} (\mathbi{\DesignVar}) \label{e:MFGPstd}
\end{gather}
\noindent where $\Kronecker_{ij}$ is the Kronecker delta function.

The multifidelity acquisition function is computed starting from the information of the multifidelity GP. Multifidelity acquisition functions include the multifidelity expected improvement (MFEI) \cite{HuangAl2006}, Multifidelity Predictive Entropy Search (MFES) \cite{ZhangAl2017}, or Multifidelity Max-Value Entropy Search \cite{TakenoAl2020}. We base our multifidelity formulation on the MFEI, which derives from the Expected improvement acquisition function: 
\begin{equation} \label{e:MFAF1}
MFEI(\mathbi{\DesignVar}, \LevFid) =  EI(\mathbi{\DesignVar}) \alpha_1(\mathbi{\DesignVar},\LevFid) \alpha_2(\mathbi{\DesignVar},\LevFid)  \alpha_3(\mathbi{\DesignVar},\LevFid)
\end{equation} 
\noindent where $\EI(\mathbi{\DesignVar})$ is the expected improvement depicted in Equation~\ref{e:EI} evaluated at the highest level of fidelity. The utility functions $\alpha_1$, $\alpha_2$ and $\alpha_3$ are defined as follows: 
\begin{equation}\label{e:MFAF3}
    \alpha_1 (\mathbi{\DesignVar}, \LevFid) = corr \left[ \ObjFun^{(\LevFid)}, \ObjFun^{(\MaxLevFid)} \right] 
\end{equation}  
\begin{equation} \label{e:MFAF4}
    \alpha_2 (\mathbi{\DesignVar}, \LevFid) = 1 - \frac{\StandDevNoise}{\sqrt{\StandDevGP^{2(\LevFid)} (\mathbi{\DesignVar}) + \StandDevNoise^{2}}}
\end{equation}
\begin{equation} \label{e:MFAF5} 
    \alpha_3 (\LevFid) = \frac{\CompCost^{(\MaxLevFid)}}{\CompCost^{(\LevFid)}}.
\end{equation}
\noindent $\alpha_1 (\mathbi{\DesignVar}, \LevFid)$ accounts the increase of the objective uncertainty as the level of fidelity decreases; it is defined as the correlation between the evaluation of the objective function at the $\LevFid$-th fidelity and the high-fidelity evaluation. $\alpha_2 (\mathbi{\DesignVar}, \LevFid)$ considers the reduction of the uncertainty on the multifidelity GP prediction after a new evaluation of the objective function. $\alpha_3 (\LevFid)$ accounts for the computational cost associated with the level of fidelity of the objective representations: it is defined as the ratio between the cost per evaluation of the high-fidelity model and the cost of the $\LevFid$-th fidelity model.

\section{Non-Myopic Multifidelity Bayesian Optimization }
\label{s: Non-myopic multifidelity active learning}

 In this section, we formalize the MFBO as a dynamic programming problem and define the two-step lookahead multifidelity acquisition function (Section~\ref{s: Two-step lookahead multifidelity acquisition function}). The solution of our acquisition function requires an approximation of its intractable formulation for which we adopt the Monte Carlo approach  (Section~\ref{s: Estimate of the two-step lookahead multifidelity acquisition function}). The lookahead strategy is then combined with a Bayesian approach to multifidelity optimization into our non-myopic MFBO scheme (Section~\ref{s: Non-myopic multifidelity algorithm}).
 


\subsection{Formulating the Two-Step Lookahead Multifidelity Acquisition Function}
\label{s: Two-step lookahead multifidelity acquisition function}

Multifidelity Bayesian optimization can be formulated as a dynamic programming (DP) problem. DP is conceived to address decision problems modeling the decision making process under uncertainty as a dynamic systems, where its dynamics characterize the sequence of decisions to achieve a specific goal \cite{Bertsekas1995, Powell2007}. Considering MFBO as a system governed by a discrete stage dynamics, the DP problem consists of (i) a dynamic system as the probabilistic surrogate model that synthesizes the multifidelity representations of the objective function, (ii) a system dynamics that describes how the surrogate model is updated when a new input is observed at a certain level of fidelity, and (iii) a goal that can be quantified with a measure of the two-step ahead improvement of the solution quality achieved by evaluating a specific input. 

DP addresses the decision making process through a policy that determine a decision given the information about the dynamic system. In the following, we aim to construct an optimal policy for MFBO that allows to select the input point together with the level of fidelity to query predicting benefits two steps ahead.


Let us consider the generic stage $\Stage$ of MFBO, and define the input $\DesignVar$ and the observation of the objective function $\NoisyObserv^{(\LevFid)}$ at the $\LevFid$-th level of fidelity, where the levels of fidelity can be in principle extended to any finite number. The system is fully characterized by a state $\State_{\Stage}$ corresponding to the training dataset $\Dataset_{\Stage}=\{\DesignVar_{\IndNumObs},\NoisyObserv^{(\LevFid_{\IndNumObs})},\LevFid_{\IndNumObs} \}_{\IndNumObs = 1}^{\NumObs}$ of $\NumObs$ observations, and a control $\Control_{\Stage} = \{\DesignVar_{\Stage+1}, \LevFid_{\Stage+1} \}$ that activates the dynamics of the system. Thus, a policy $\Policy_{\Stage}$ is defined as a function that maps the state $\State_{\Stage}$ to a control $\Control_{\Stage}=\Policy_{\Stage}(\State_{\Stage})$.  

Given the state and the control of the system, we introduce the disturbances to represent a simulated value of the objective function at $\{\DesignVar_{\Stage+1},\LevFid_{\Stage+1} \}$, defined as a random variable $\Disturbances_{\Stage}^{(\LevFid)} \sim \NorDist (  \MeanFunGP_{\Stage}^{(\LevFid)} (\DesignVar_{\Stage+1}), \StandDevGP_{\Stage}^{2(\LevFid)} (\DesignVar_{\Stage+1}))$ characterized with the mean (Equation~\eqref{e:MFGPmean}) and variance (Equation~\eqref{e:MFGPstd}) of the multifidelity Gaussian process. At the new stage $\Stage+1$, the system evolves to a new state $\State_{\Stage+1}$ following its dynamics, corresponding to the augmented dataset $\Dataset_{\Stage+1} = \Dataset_{\Stage} \cup \{\DesignVar_{\Stage+1},\NoisyObserv^{(\LevFid_{\Stage+1})}, \LevFid_{\Stage+1} \}$: 
\begin{equation} \label{e:2LookMFAF1}
    \Dataset_{\Stage+1}=\SysDyn(\DesignVar_{\Stage+1},\NoisyObserv^{(\LevFid_{\Stage+1})}, \LevFid_{\Stage+1}, \Dataset_{\Stage})
\end{equation}
The disturbances $\Disturbances_{\Stage+1}$ are then characterized using the multifidelity Gaussian process conditioned on $\Dataset_{\Stage+1}$. We define a stage reward function to measure the benefits of applying the control $\Control_{\Stage}$ to a state $\State_{\Stage}$ subject to the disturbances $\Disturbances_{\Stage}$. For the MFBO dynamic system, we formulate the stage reward function as the reduction of the objective function achieved at the stage $\Stage+1$ with respect to $\Stage$: 
\begin{equation}\label{e:2LookMFAF2}
    \StageReward_{\Stage}(\DesignVar_{\Stage+1},\NoisyObserv^{(\LevFid_{\Stage+1})}, \LevFid_{\Stage+1}, \Dataset_{\Stage}) = (\ObjFun_{\Stage}^{*(\MaxLevFid)} - \ObjFun_{\Stage+1}^{(\MaxLevFid)})^{+}
\end{equation}
\noindent where $\ObjFun_{\Stage}^{*(\MaxLevFid)}$ is the minimum value of the objective function at $\Stage$ evaluated at the highest level of fidelity. Thus, we can formulate the two-step lookahead multifidelity acquisition function at a generic stage $\Stage$ as the expected reward: 
\begin{equation}\label{e:2LookMFAF3}
\begin{aligned}
    \AF^{\Policy}_{\Stage} (\DesignVar_{\Stage+1}, \LevFid_{\Stage+1}, \Dataset_{\Stage})  =    
      \Expectation [ \StageReward_{\Stage}(\DesignVar_{\Stage+1},\NoisyObserv^{(\LevFid_{\Stage+1})}, \LevFid_{\Stage+1}, \Dataset_{\Stage})
      + \ExpectedReward_{\Stage+1}(\SysDyn(\DesignVar_{\Stage+1},\NoisyObserv^{(\LevFid_{\Stage+1})}, \LevFid_{\Stage+1}, \Dataset_{\Stage}))]  
\end{aligned}
\end{equation}
\noindent where the expectation is taken with respect to the disturbances, $\Expectation \left[ \StageReward_{\Stage}(\cdot) \right] = \MFEI (\DesignVar_{\Stage+1}, \LevFid_{\Stage+1})$ is the multifidelity expected improvement (Equation~\eqref{e:MFAF1}), and $\ExpectedReward_{\Stage+1} (\SysDyn( \cdot))$ is the long-term expected reward. 

Following the DP principle, the objective is to maximize the long term reward defining an optimal policy $\Policy^{*}_{\Stage}$. For the case of our two-step lookahead multifidelity acquisition function, the optimal policy is the one that identifies the optimal input variable and the associated level of fidelity to query at the second future step to minimize the cumulative expected loss. Thus, we define the long term reward $\ExpectedReward_{\Stage+1}$ as the maximum of the multifidelity expected improvement conditioned on the training set $\Dataset_{\Stage+1}$:   
\begin{equation}\label{e:2LookMFAF4}
    \ExpectedReward_{\Stage+1} = \max(\MFEI(\DesignVar_{\Stage+2}, \LevFid_{\Stage+2}))
\end{equation}
Combining Equation~\eqref{e:2LookMFAF3} with Equation~\eqref{e:2LookMFAF4}, we formalize the two-step lookahead multifidelity acquisition function:
\begin{equation}\label{e:2LookMFAF5}
\begin{aligned}
    \AF^{\Policy^{*}}_{\Stage}(\DesignVar_{\Stage+1},\LevFid_{\Stage+1}&,\Dataset_{\Stage}, \Dataset_{\Stage+1})  = 
    \MFEI (\DesignVar_{\Stage+1}, \LevFid_{\Stage+1}) +  \Expectation \left[\max(\MFEI(\DesignVar_{\Stage+2},\LevFid_{\Stage+2}))\right]
\end{aligned}
\end{equation}

\subsection{Estimating the Two-Step Lookahead Multifidelity Acquisition Function}
\label{s: Estimate of the two-step lookahead multifidelity acquisition function}

The solution of the two-step lookahead multifidelity acquisition function (Equation~\eqref{e:2LookMFAF5}) cannot be computed in closed-form, as requires to evaluate the nested maximizations and expectations. 

We adopt the Monte Carlo approach to estimate $\AF_{\Stage}^{\Policy^{*}}$, using the reparameterization strategy proposed by Wilson et al. \cite{WilsonAl2018} to formulate the value of the objective function at the first step ahead with $\LevFid$-th level of fidelity: 
\begin{equation} \label{e:MC1}
    \ObjFun^{(\LevFid)} \left(\DesignVar_{\Stage+1} \right) = \MeanFunGP^{(\LevFid)}_{\Stage} + \mathbi{\CholDec}^{(\LevFid)}_{\Stage} \left(\DesignVar_{\Stage+1} \right) \NorVar
\end{equation}
\noindent where $\mathbi{\CholDec}^{(\LevFid)}_{\Stage}$ is the Cholesky decomposition of the covariance matrix $\mathbi{\KernelMatrix}_{\Stage}$, and $\NorVar$ is an independent standard normal random variable. Then, we use Equation~\eqref{e:MC1} to compute the mean and variance of the multifidelity GP at the step $\Stage+1$ for the generic input $\DesignVar$: 
\begin{equation} \label{e:MC2}
\MeanFunGP^{(\LevFid)}_{\Stage+1}(\mathbi{\DesignVar}) = \MeanFunGP^{(\LevFid)}_{\Stage}(\mathbi{\DesignVar}) + \mathbi{\AuxMat}^{(\LevFid)}_{\Stage} \left( \mathbi{\DesignVar}\right) \NorVar  
\end{equation}
\begin{equation} \label{e:MC3}
\StandDevGP^{(\LevFid)}_{\Stage+1}(\mathbi{\DesignVar}) = \StandDevGP^{(\LevFid)}_{\Stage}(\mathbi{\DesignVar}) - \mathbi{\AuxMat}^{(\LevFid)}_{\Stage} \left( \mathbi{\DesignVar}\right) \mathbi{\AuxMat}^{(\LevFid)}_{\Stage} \left( \mathbi{\DesignVar}\right)^T
\end{equation}
\noindent where $\mathbi{\AuxMat}^{(\LevFid)}_{\Stage} \left( \mathbi{\DesignVar}\right) = \CovFunGP^{(\LevFid)}_{\Stage} \left( \mathbi{\DesignVar}\right) \mathbi{\CholDec}^{(\LevFid)-1}_{\Stage}(\mathbi{\DesignVar})$. 

Equation~\eqref{e:MC2} and Equation~\eqref{e:MC3} are used to estimate the multifidelity expected improvement at the step $\Stage+1$: 
\begin{equation}\label{e:MC4}
\begin{aligned}
   \MFEI(\DesignVar_{\Stage+2}, \LevFid_{\Stage+2}) \sim \widehat{\MFEI}(\DesignVar_{\Stage+2},\LevFid_{\Stage+2},\NorVar) 
\end{aligned}
\end{equation}

To approximate the expectation term of Equation~\eqref{e:2LookMFAF5}, we sample the random variable $\NorVar$ and compute $\Expectation[\widehat{\AF}^{\Policy^{*}}_{\Stage}(\DesignVar_{\Stage+1},\LevFid_{\Stage+1}, \NorVar)]$ averaging over many realizations of the two-step lookahead multifidelity acquisition $\widehat{\AF}_{\Stage}^{\Policy^{*}}$ evaluated using Equation~\eqref{e:MC4}.

Although the need of numerous evaluations of $\widehat{\AF}^{\Policy^{*}}_{\Stage}$, we emphasize that our Monte Carlo approach requires the evaluation of Equation~\eqref{e:MC2} and Equation~\eqref{e:MC3} which are inexpensive to compute. Thus, the computational cost associated with the approximation of the two-step lookahead multifidelity acquisition function is negligible if compared with the cost of real-world optimization problems in science and engineering applications, where usually the evaluation of the objective function requires the query of expensive physics-based models.

\subsection{Non-Myopic Multifidelity Optimization Algorithm}
\label{s: Non-myopic multifidelity algorithm}
 
\begin{algorithm}[t!] 
\caption{Non-myopic multifidelity Bayesian optimization}
\label{a:OptimizationAlgorithm}
\begin{algorithmic}[1]

\REQUIRE  Feasible set $\DesignSpace \in \mathbb{R}^{\Dim}$, multifidelity objective function $\ObjFun^{(\LevFid)}(\DesignVar)$ and the multifidelity Gaussian process prior $GP(0, \CovFunGP^{(\LevFid)} (\DesignVar, \DesignVar')) $

\ENSURE $\MinObjFun = \min \ObjFun(\DesignVar)$

\STATE $\Dataset_0 \leftarrow \{ \DesignVar_{\IndNumObs},\NoisyObserv^{(\LevFid_\IndNumObs)},\LevFid_{\IndNumObs} \}_{\IndNumObs=1}^{\NumObs_0}$ collect initial observations

\STATE $\MeanFunGP^{(\LevFid)}_0, \StandDevGP^{2(\LevFid)}_0\ \leftarrow$ learn the initial multifidelity GP 

\STATE $\IterOpt \leftarrow 1$

\REPEAT 
       \STATE {Load $\DesignVar_{\IterOpt}$ and associated $\LevFid_{\IterOpt}$}

       \STATE Evaluate $\NoisyObserv^{(\LevFid_\IterOpt)}(\DesignVar_{\IterOpt})$  
       

       \STATE $\Dataset_{\IterOpt} \leftarrow \Dataset_{\IterOpt-1} \cup \{ \DesignVar_{\IterOpt},\NoisyObserv^{(\LevFid_\IterOpt)},\LevFid_{\IterOpt} \}$

       \STATE $\MeanFunGP^{(\LevFid)}_\IterOpt, \StandDevGP^{2(\LevFid)}_\IterOpt \leftarrow$ update the multifidelity GP 

       \STATE $\Stage \leftarrow \IterOpt$
       
       \STATE Compute $\MFEI (\DesignVar_{\Stage+1}, \LevFid_{\Stage+1})$

       \FOR {$\MCiter \leftarrow 1, \NumObs_{MC}$} 
            
            \STATE $\NorVar_{\MCiter} \leftarrow \NorDist(0,1)$
            
            \STATE $\MeanFunGP^{(\LevFid)}_{\Stage+1}, \StandDevGP^{2(\LevFid)}_{\Stage+1} \leftarrow$ estimate the multifidelity GP 
            
            \STATE Compute $\widehat{\MFEI}(\DesignVar_{\Stage+2},\LevFid_{\Stage+2},\NorVar_{\MCiter}) $
       
       \ENDFOR

\STATE \textbf{return} $\widehat{\mathbi{\AF}}^{\Policy^{*}}_{\Stage} = \{\widehat{\AF}^{\Policy^{*}}_{\Stage_\MCiter} \}_{\MCiter=1}^{\NumObs_{MC}}$

\STATE $\AF^{\Policy^{*}}_{\IterOpt} = \Expectation \left[ \widehat{\mathbi{\AF}}^{\Policy^{*}}_{\Stage}\right] $

\STATE $\left[ \DesignVar_{\IterOpt+1}, \LevFid_{\IterOpt+1} \right] \leftarrow \max (\AF_{\IterOpt}^{\Policy^{*}})$

\STATE $\IterOpt+1 \leftarrow \IterOpt$

\UNTIL {$\Budget_{\IterOpt} \leq \Budget_{max}$}


\STATE \textbf{return} $\MinDesignVar$ that minimize $\ObjFun(\DesignVar)$ over $\Dataset_{\IterOpt}$ 

\end{algorithmic}

\end{algorithm}

Algorithm~\ref{a:OptimizationAlgorithm} details the steps of our non-myopic multifidelity optimization scheme. The optimization starts with the definition of the $\Dim$-dimensional input space $\DesignSpace \in \mathbb{R}^\Dim$ assembled through a Latin Hypercube design of experiments \cite{Mckay1992}, together with the objective function at different levels of fidelity $\ObjFun^{(\LevFid)}$, and the multifidelity Gaussian process prior $GP(0,\CovFunGP^{(\LevFid)}(\DesignVar,\DesignVar'))$. We draw an initial subset of feasible input points $\{\DesignVar_{\IndNumObs}\}_{\IndNumObs=1}^{\NumObs_0}$ and associated levels of fidelity $\{ \LevFid_{\IndNumObs}\}_{\IndNumObs=1}^{\NumObs_0}$, where $\NumObs_0$ is the number of points sampled at the beginning of the optimization. Contextually, we collect observations of the objective function $\{ \NoisyObserv^{(\LevFid_\IndNumObs)} \}_{\IndNumObs=1}^{\NumObs_0}$ and determine the initial state $\Dataset_0 = \{ \DesignVar_{\IndNumObs},\NoisyObserv^{(\LevFid_\IndNumObs)},\LevFid_{\IndNumObs} \}_{\IndNumObs=1}^{\NumObs_0}$. The multifidelity GP prior together with the initial state $\Dataset_0$ induce the multifidelity Gaussian process posterior with mean $\MeanFunGP^{(\LevFid)}$ (Equation~\eqref{e:MFGPmean}) and variance $\StandDevGP^{2(\LevFid)}$ (Equation~\eqref{e:MFGPstd}), representing the first surrogate model of the objective function. 

Let us now consider the generic iteration $\IterOpt$ of the optimization search. The algorithm selects the input $\DesignVar_{\IterOpt}$ along with the associated level of fidelity $\LevFid_{\IterOpt}$ at the previous step $\IterOpt-1$, and computes the value of the objective function $\NoisyObserv^{(\LevFid_{\IterOpt})}(\DesignVar_{\IterOpt})$. This information is used to augment the state $\Dataset_{\IterOpt} = \Dataset_{\IterOpt-1} \cup \{\DesignVar_{\IterOpt},\NoisyObserv^{(\LevFid_\IterOpt)},\LevFid_{\IterOpt}\}$ and update the multifidelity Gaussian process conditioned on $\Dataset_{\IterOpt}$. 

At the same iteration $\IterOpt$, we compute the two-step lookahead multifidelity acquisition function to determine the next input $\DesignVar_{\IterOpt+1}$ to evaluate and the associated level of fidelity $\LevFid_{\IterOpt+1}$. We indicate with $\Stage=\IterOpt$ the current step of the optimization and with $\Stage+1$ and $\Stage+2$ the first and second step ahead, respectively. The first term of the acquisition function  (Equation~\eqref{e:2LookMFAF5}) is computed on the current state $\Dataset_{\Stage}=\Dataset_{\IterOpt}$, while the second term is estimated using the Monte-Carlo approach discussed in Section~\ref{s: Estimate of the two-step lookahead multifidelity acquisition function}. Considering the $\MCiter$-th Monte Carlo simulation, we sample independently a random variable $\NorVar_{\MCiter}$ normally distributed, and estimate the mean $\MeanFunGP^{(\LevFid)}_{\State+1}$ and variance $\StandDevGP^{2(\LevFid)}_{\State+1}$ of the multifidelity GP computing Equation~\eqref{e:MC2} and Equation~\eqref{e:MC3}. Then, we evaluate the second term of the acquisition function through Equation~\eqref{e:MC4} and compute $\widehat{\AF}^{\Policy^{*}}_{\Stage_\MCiter}$. The Monte Carlo algorithm iterates till a maximum number of simulations $\NumObs_{MC}$ is reached and returns the realizations of the approximated acquisition function $\{\widehat{\AF}^{\Policy^{*}}_{\Stage_\MCiter} \}_{\MCiter=1}^{\NumObs_{MC}}$. This allows to estimate the two-step multifidelity acquisition function $\AF_{\IterOpt}$ as the expectation taken over the Monte Carlo simulations. 

Then, similarly to the multifidelity Bayesian optimization (Section~\ref{s: Multifidelity Bayesian optimization}), we maximize $\AF^{\Policy^{*}}_{\IterOpt}$ and determine the next input to evaluate $\DesignVar_{\IterOpt+1}$ and the associated level of fidelity $\LevFid_{\IterOpt+1}$. The optimization is iterated until a maximum computational budget $\Budget_{\IterOpt} = \Budget_{max}$ is reached, where $\Budget_{\IterOpt}$ is the cumulative computational cost expended until iteration $\IterOpt$.

\section{Experiments}
\label{s: Numerical experiments}

In this section, we illustrate and discuss the proposed non-myopic MFBO in comparison with MFBO over a set of well accepted benchmark problems, specifically conceived to stress-test multifidelity methods. As the standard MFBO algorithm, we implemented the method based on multifidelity expected improvement proposed by Huang et al. \cite{HuangAl2006}. 

For all the results presented in this section, we impose the same initial settings for both the non-myopic and the standard MFBO algorithm. We define the number of points $\IndNumObs_0^{(\LevFid)}$ for each $\LevFid$-th fidelity level that characterize the initial dataset $\Dataset_{0}$, together with the maximum computational budget $\Budget_{max}$ quantified as the sum of the computational costs associated with observations of the objective function at the $\LevFid$-th level of fidelity. The initial observations of the objective function at different levels of fidelity are determined through a Latin Hypercube Sampling \cite{Mckay1992} and are used to compute the first surrogate model as depicted in Section~\ref{s: Non-myopic multifidelity algorithm}. We use the square exponential kernels for all the GP covariances, and optimize the hyperparameters of the kernel and mean functions of the GP via Maximum Likelihood Estimation \cite{ForresterAl2008}.

The set of multifidelity benchmark functions we used to validate our methodology is proposed by Mainini et al. \cite{MaininiAl2022} and includes the Forrester function, the Rosenbrock function, the Rastrigin function shifted and rotated and a mass spring system optimization problem. The multifidelity Borehole function test-case \cite{XiongAl2013} is also included to further expand the experimentation. This set of functions is specifically selected to emulate mathematical characteristics that are frequent in physics-based optimization problems, keeping contained the computational expense associated with their evaluation. Table~\ref{t:Settings} summarizes the experimental setup for each test function.

\begin{table}[h]
  \caption{Summary of the experiments setup}
  \label{t:Settings}
  \centering
  \begin{tabular}{llllll}
    \toprule
    Function & $\IndNumObs_0^{(2)}$ & $\IndNumObs_0^{(1)}$ & $\Budget_{max}$ & $\CompCost^{(2)}$  & $\CompCost^{(1)}$  \\
    \midrule
    Forrester & 2 & 5 &  100 & 1 &  0.05 \\
Rosenbrock 2D & 5 & 10 &  200 & 1 &  0.5 \\
Rosenbrock 5D & 15 & 30 &  500 & 1 &  0.5 \\
Rosenbrock 10D & 50 & 250 &  1000 & 1 &  0.5 \\
Rastrigin  & 10 & 20 &  200 & 1 &  0.0039 \\
Mass Spring & 4 & 10 &  400 & 1 &  1/60 \\
Borehole & 100 & 500 &  800 & 1 &  0.5 \\
    \bottomrule
  \end{tabular}
\end{table}

\subsection{Results and Discussion}
\label{s:results}

In the following, the results are observed and discussed in terms of difference error normalized in the input space: \begin{equation} \label{e:error}
    \Error = \frac{\ObjFun(\DesignVar^{*}_{\IterOpt}) - \MinObjFun}{\ObjFun_{max} - \MinObjFun}
\end{equation}
\noindent where $\ObjFun(\DesignVar^{*}_{\IterOpt})$ is the best value of the objective function determined by the algorithm at the $\IterOpt$-th iteration, $\MinObjFun$ is the analytical solution of Equation~\eqref{e:OptPrb} and $\ObjFun_{max} = \max_{\DesignVar \in \DesignSpace} \ObjFun(\DesignVar)$. The error $\Error$ is plotted as a function of the computational Budget $\Budget$ to compare the optimization results achieved by the algorithms. To quantify and compensate the influence of the random design of experiments, we carried out the tests by running 5 trials considering a random initialization for each benchmark problem, and reported the result in terms of median values of the error together with the values falling in the interval between the 25-th and 75-th percentiles.

\begin{figure*}[ht]
    \centering
     \subfigure[Forrester]{%
        \includegraphics[width=0.3\linewidth,trim=220 0 245 0, clip]{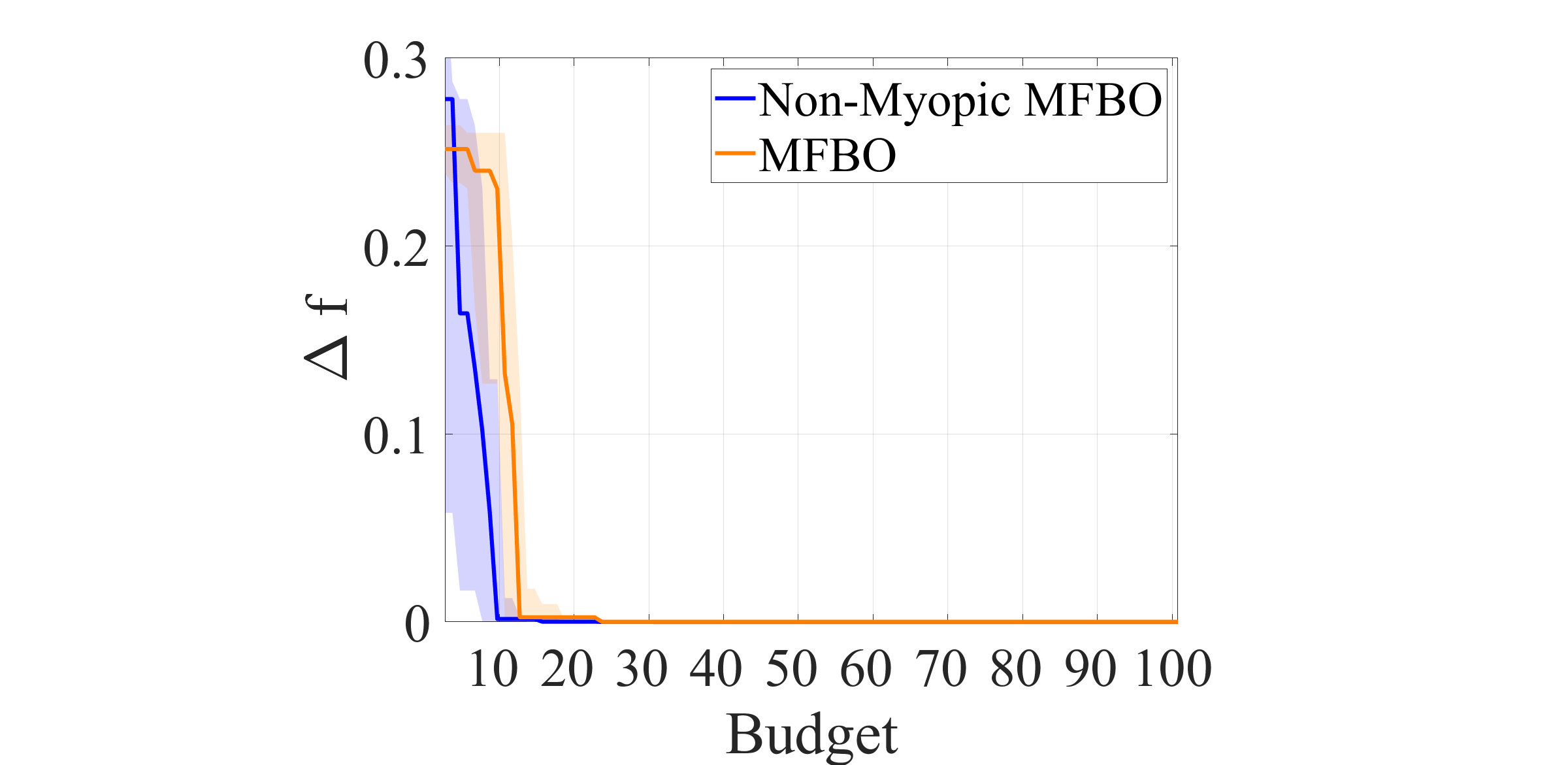} 
        \label{fig:Forrester}}
     \subfigure[Rastrigin shifted and rotated]{
        \includegraphics[width=0.3105\linewidth,trim=200 0 244 0, clip]{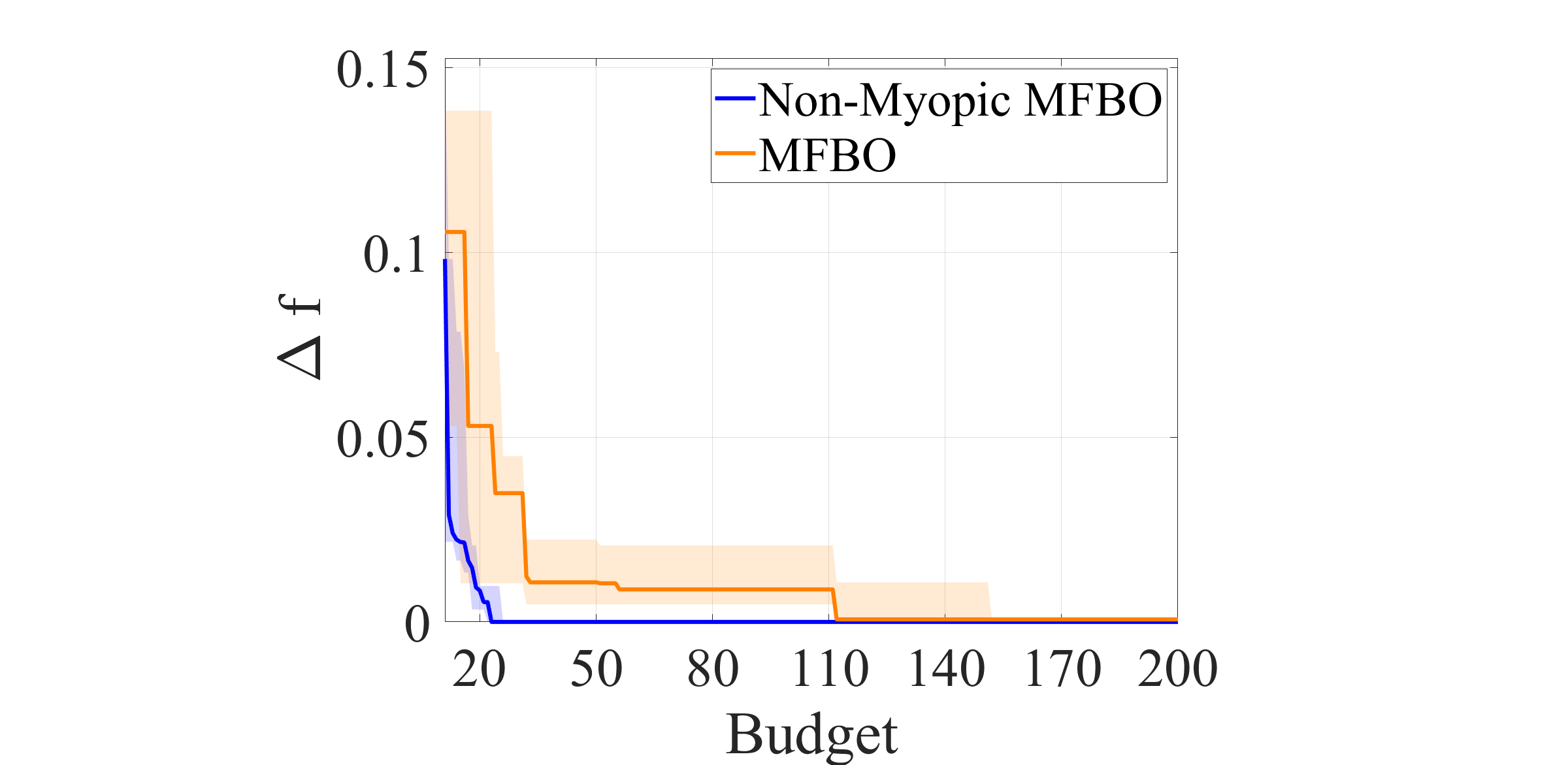}
        \label{fig:Rastrigin}}
        
     \subfigure[Rosenbrock 2D]{%
        \includegraphics[width=0.31\linewidth,trim=200 0 243 0, clip]{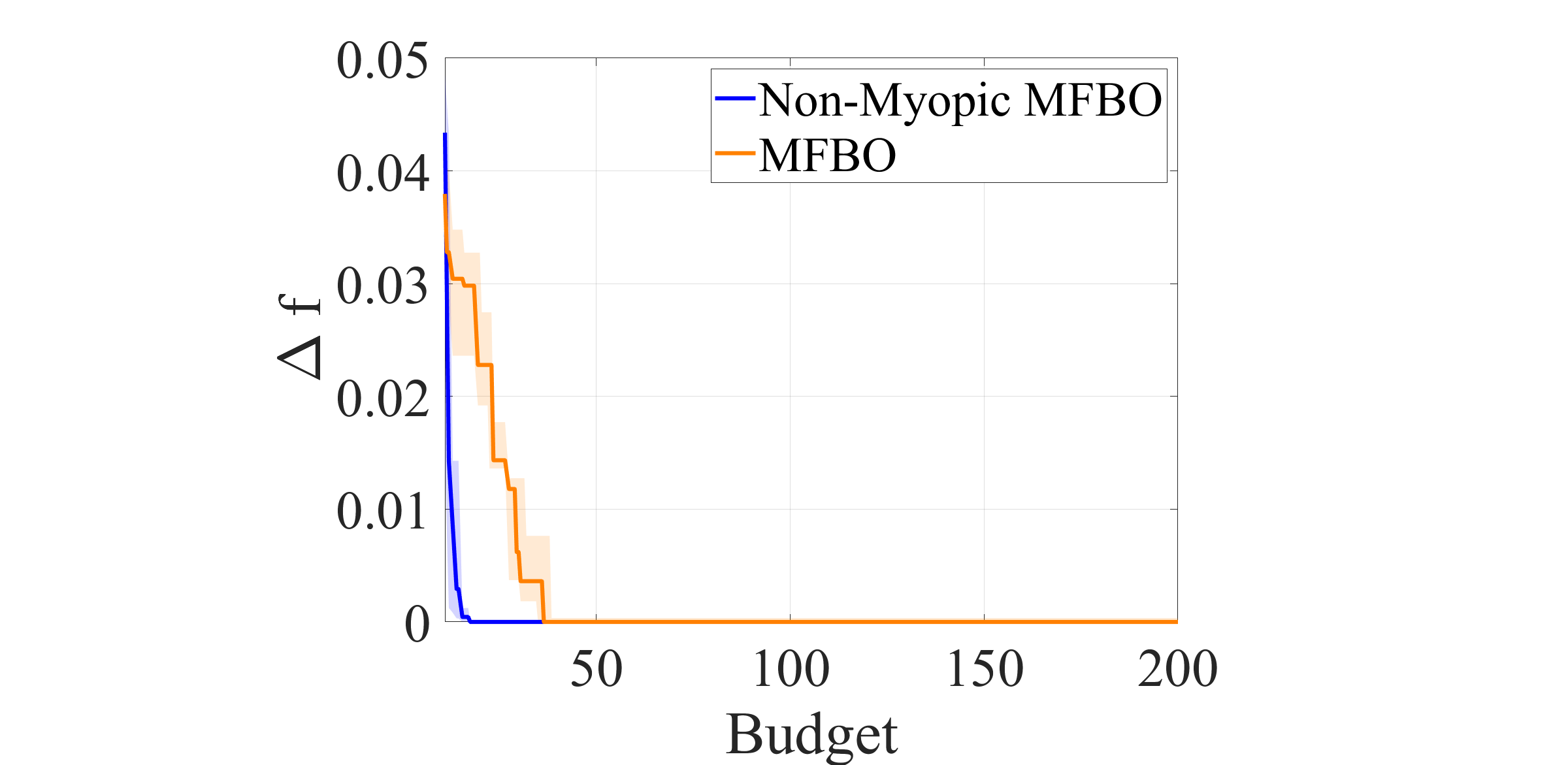}}
        \label{fig:Rosenbrock2D}  
     \subfigure[Rosenbrock 5D]{%
        \includegraphics[width=0.31\linewidth,trim=200 0 243 0, clip]{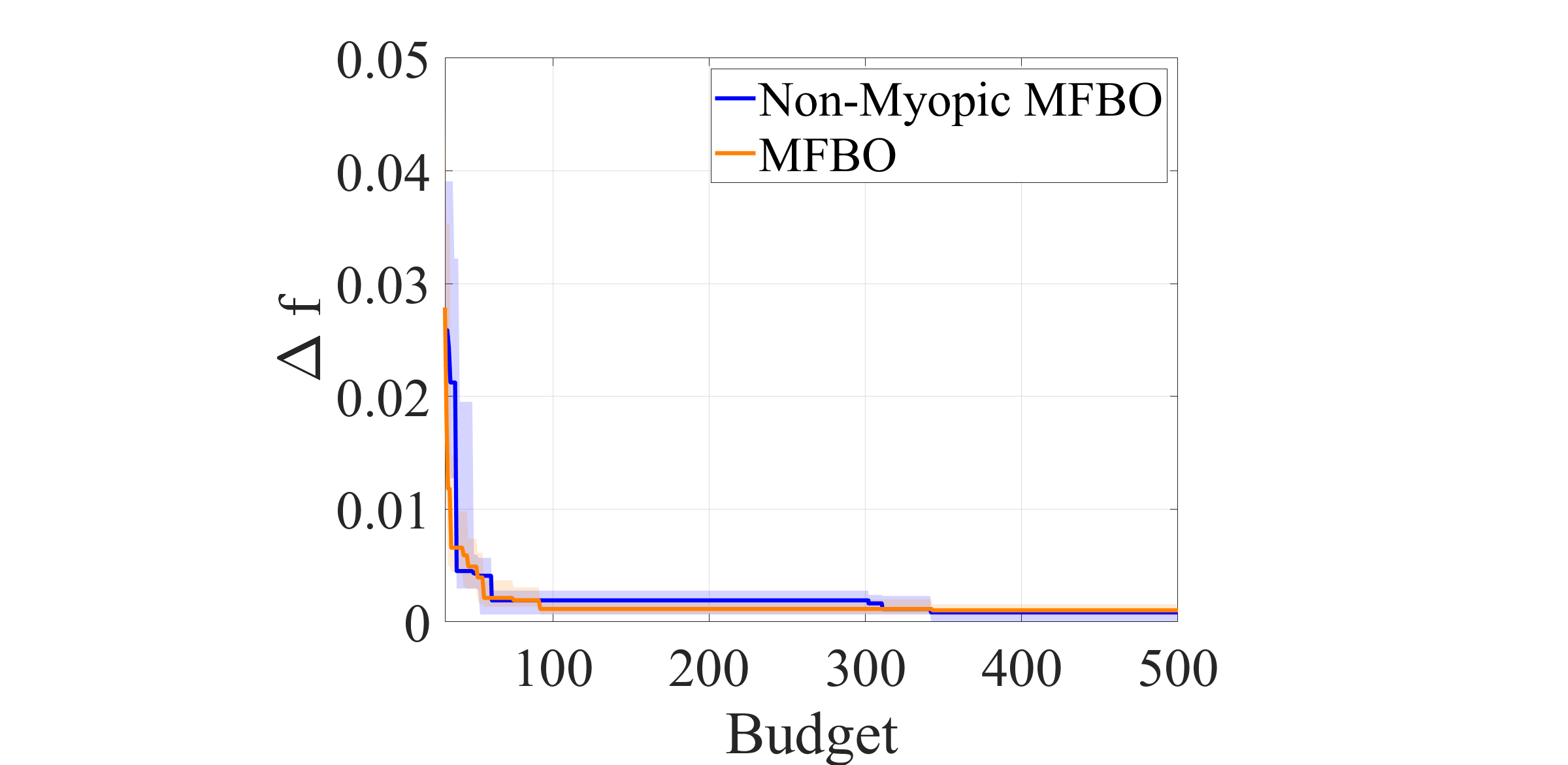}}
        \label{fig:Rosenbrock5D}
     \subfigure[Rosenbrock 10D]{
        \includegraphics[width=0.325\linewidth,trim=180 0 230 0, clip]{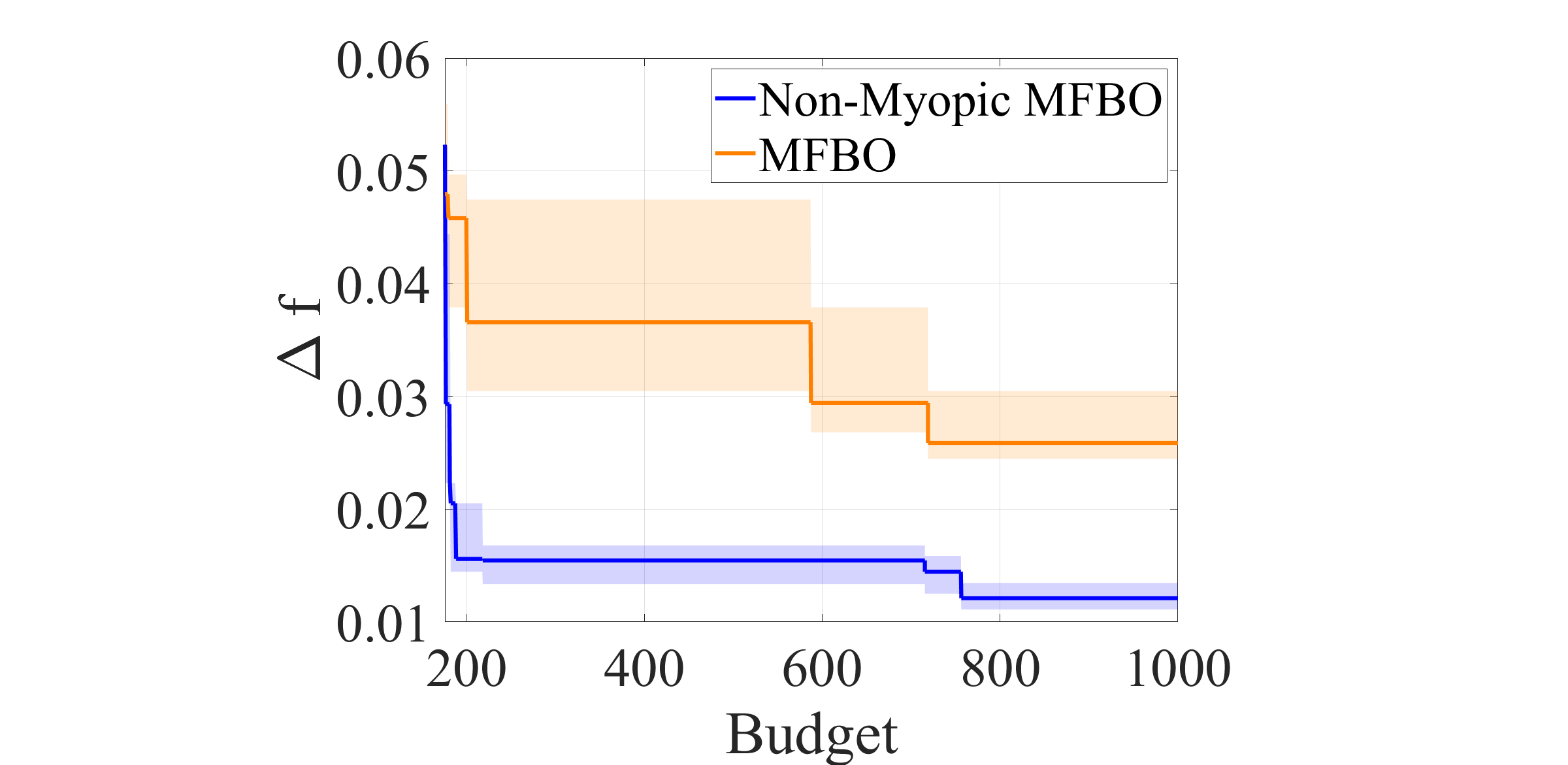}
        \label{fig:Rosenbrock10D}}
    \subfigure[Mass spring system]{%
        \includegraphics[width=0.31\linewidth,trim=220 0 240 0, clip]{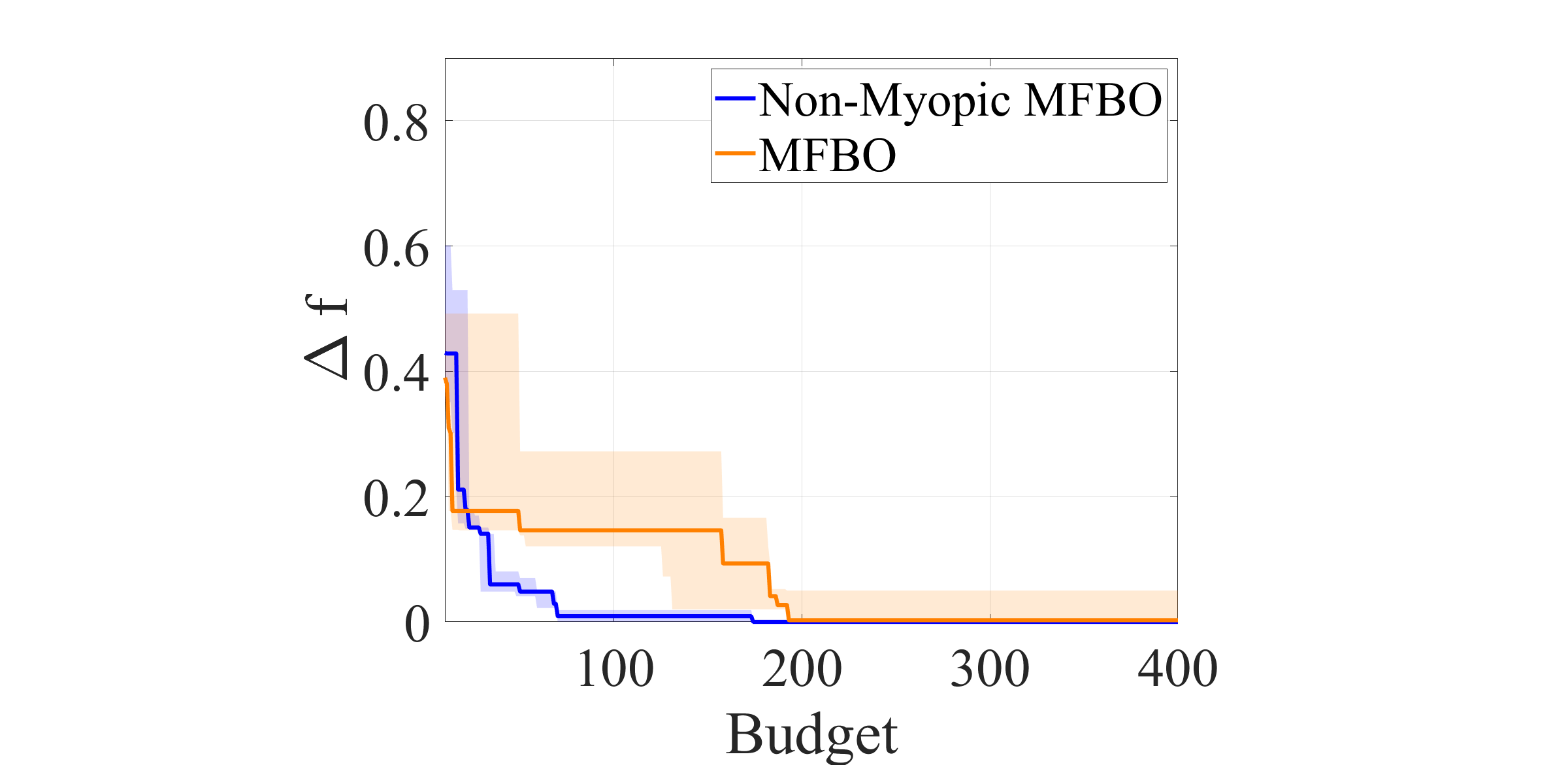}}
        \label{fig:MK4}
     \subfigure[Borehole]{
        \includegraphics[width=0.32\linewidth,trim=200 0 240 0, clip]{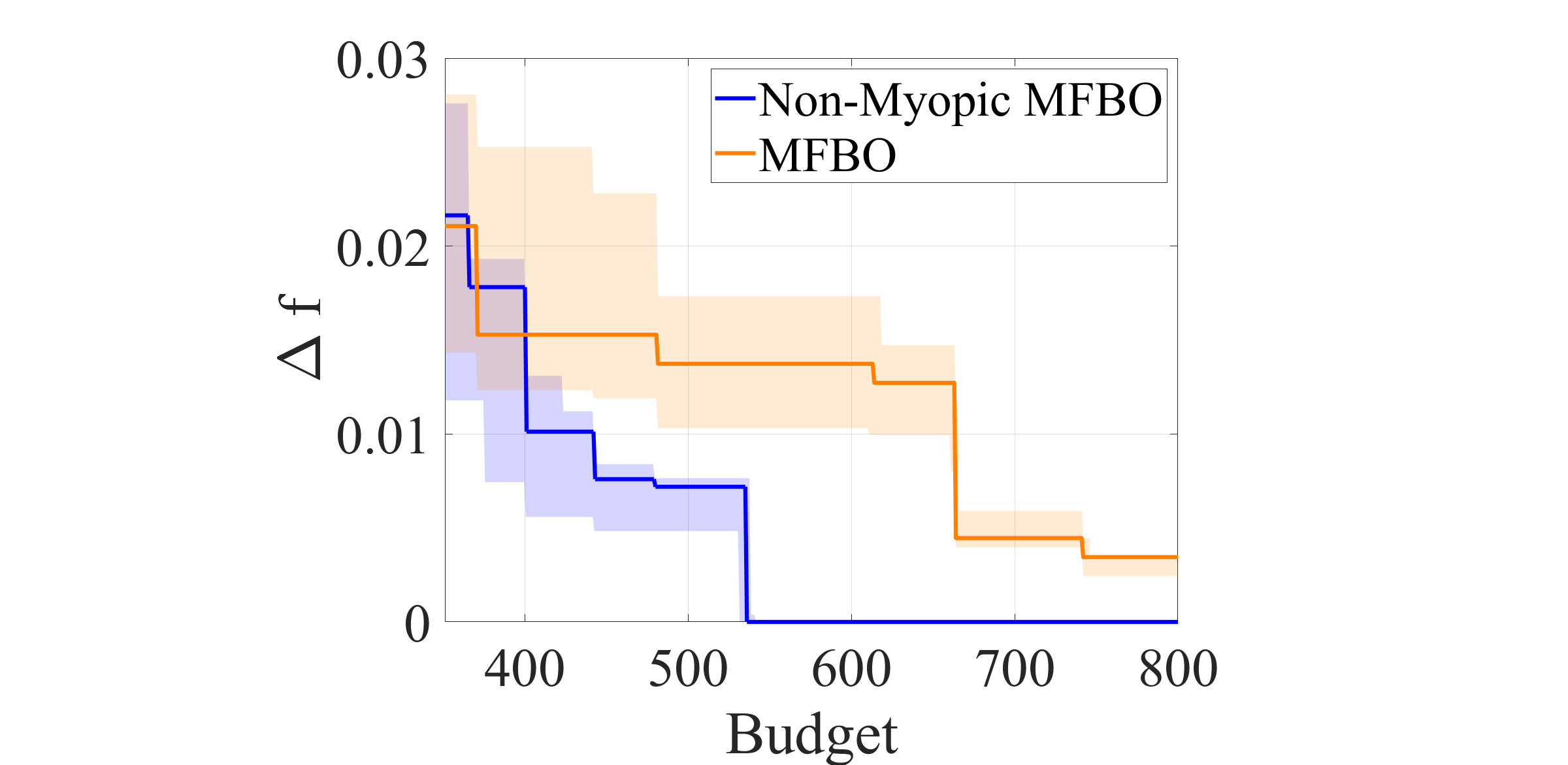}
        \label{fig:Borehole}}
    \caption{Performances of the Non-Myopic MFBO compared to the standard MFBO.}\label{fig:Results}
\end{figure*}

Figure~\ref{fig:Results} summarizes the results we obtained from the experiments discussed in Section~\ref{s: Numerical experiments}. In all the benchmark problems, the non-myopic multifidelity Bayesian optimization (Non-Myopic MFBO) outperforms the standard multifidelity Bayesian framework (MFBO) in terms of the reduction of the error values, with a much smaller or comparable computational cost. 

Indeed, considering the optimization of the Forrester function (figure 1(a)), our algorithm permits to determine the optimal solution with a significant reduction of the expended budget if compared with the baseline MFBO algorithm. 

A significant performance of our non-myopic framework can be observed in the optimization of the Rastrigin function (figure 1(b)), where the step-like behaviour of the error occurs when the optimizer falls in a local minimum of the multimodal function, leading to a constant error until a new minimum is observed. This behavior is emphasized for the MFBO while is almost not appreciable for the non-myopic MFBO, suggesting that the lookahead property allows to handle objective functions with a pronounced multimodal nature. A possible explanation for this finding might be attributed to an efficient form of exploration, where the observation of high-uncertainty regions of the input domain improves future evaluations, leading to an effective exploitation toward the optimum. This is in agreement with what observed by Ginsbourger and Le Riche \cite{Ginsbourger&LeRiche2010} for their non-myopic single-fidelity optimization algorithm.

The experiments on the Rosenbrock function are conducted for input spaces of various dimensionality ($\Dim=2,5,10$). For $\Dim=2$ (figure 1(c)), the non-myopic MFBO outperforms the standard MFBO achieving a much smaller error with a fraction of the cost required by the standard framework. As the dimension increases to $\Dim=5$ (figure 1(d)), although the MFBO is able to reduce the error in the initial stage, it eventually converges to a sub-optimal solution compared with the non-myopic strategy. At $\Dim=10$ (figure 1(e)), the algorithms are not capable to reach the optimum with the allocated budget. This undesirable performance is due to the difficulties of the GPs to accurately model high dimensional functions. After the initial stages, the error decreases only with large computational expense for both the non-myopic and the standard MFBO schemes. We observed that the algorithms here mostly query the low-fidelity model to contain the computational expense during a secondary exploration phase.

In addition, the non-myopic MFBO reaches the minimum function values with a smaller cost than the competing standard MFBO for both the optimization problem of the mass spring system (figure 1(f)) and of the Borehole function (figure 1(g)), consistently with the results obtained for the other benchmark functions.

The results obtained for all the benchmark problems reveal that the non-myopic multifidelity strategy leads to overall better performance with respect to the standard multifidelity setting, demonstrating the ability to efficiently explore the input space with a contained budget and take advantages from the acquired information to locally exploit toward the optimum.

All the experiments presented in this paper are obtained running each test on a single core of a desktop PC with Intel Core i7-8700 (3.2 GHz) and 32 GB of RAM. The computational time required to compute a single iteration of the proposed non-myopic MFBO algorithm ranges from seconds to minutes.

\section{Conclusions}
\label{s: Conclusion}
In this paper, we proposed a non-myopic multifidelity Bayesian optimization framework based on two-step lookahead multifidelity acquisition function. Our algorithm allows to improve the solution of the optimization by unlocking the informative gain acquired from future steps of the optimization, reducing the total computational expense. The non-myopic framework can handle an unlimited finite number of fidelities levels, ensuring its applicability to a variety of real-world optimization problems. We evaluated the performances of our strategy against a standard multifidelity Bayesian framework on several benchmark optimization problems, revealing that the lookahead property allows to outperform it.

\section*{Acknowledgments}
The work has been partly conducted under the Visiting Professor Program of Politecnico di Torino. Additional acknowledgments to the Univeristy’s Doctoral Scholarship supporting Mr. Francesco Di Fiore. The authors thank Prof. Paolo Maggiore at Politecnico di Torino for the support.

\section*{References}
\renewcommand{\bibsection}{}
\bibliographystyle{unsrt}
\bibliography{bibliography}

\begin{thebibliography}{10}

\bibitem{MockusAl1978}
Jonas Mockus, Vytautas Tiesis, and Antanas Zilinskas.
\newblock The application of bayesian methods for seeking the extremum.
\newblock {\em Towards global optimization}, 2(117-129):2, 1978.

\bibitem{SnoekAl2012}
Jasper Snoek, Hugo Larochelle, and Ryan~P Adams.
\newblock Practical bayesian optimization of machine learning algorithms.
\newblock {\em Advances in neural information processing systems}, 25, 2012.

\bibitem{Kennedy&OHagan2000}
Marc~C Kennedy and Anthony O'Hagan.
\newblock Predicting the output from a complex computer code when fast
  approximations are available.
\newblock {\em Biometrika}, 87(1):1--13, 2000.

\bibitem{ForresterAl2007}
Alexander~IJ Forrester, Andr{\'a}s S{\'o}bester, and Andy~J Keane.
\newblock Multi-fidelity optimization via surrogate modelling.
\newblock {\em Proceedings of the royal society a: mathematical, physical and
  engineering sciences}, 463(2088):3251--3269, 2007.

\bibitem{PeherstorferAl2018}
Benjamin Peherstorfer, Karen Willcox, and Max Gunzburger.
\newblock Survey of multifidelity methods in uncertainty propagation,
  inference, and optimization.
\newblock {\em Siam Review}, 60(3):550--591, 2018.

\bibitem{BeranAl2020}
Philip~S Beran, Dean Bryson, Andrew~S Thelen, Matteo Diez, and Andrea Serani.
\newblock Comparison of multi-fidelity approaches for military vehicle design.
\newblock In {\em AIAA AVIATION 2020 FORUM}, page 3158, 2020.

\bibitem{HuangAl2006}
Deng Huang, Theodore~T Allen, William~I Notz, and R~Allen Miller.
\newblock Sequential kriging optimization using multiple-fidelity evaluations.
\newblock {\em Structural and Multidisciplinary Optimization}, 32(5):369--382,
  2006.

\bibitem{KandasamyAl2016}
Kirthevasan Kandasamy, Gautam Dasarathy, Junier Oliva, Jeff Schneider, and
  Barnab{\'a}s P{\'o}czos.
\newblock Gaussian process optimisation with multi-fidelity evaluations.
\newblock In {\em Proceedings of the 30th/International Conference on Advances
  in Neural Information Processing Systems (NIPS’30)}, 2016.

\bibitem{ZhangAl2017}
Yehong Zhang, Trong~Nghia Hoang, Bryan Kian~Hsiang Low, and Mohan Kankanhalli.
\newblock Information-based multi-fidelity bayesian optimization.
\newblock In {\em NIPS Workshop on Bayesian Optimization}, 2017.

\bibitem{TakenoAl2020}
Shion Takeno, Hitoshi Fukuoka, Yuhki Tsukada, Toshiyuki Koyama, Motoki Shiga,
  Ichiro Takeuchi, and Masayuki Karasuyama.
\newblock Multi-fidelity bayesian optimization with max-value entropy search
  and its parallelization.
\newblock In {\em International Conference on Machine Learning}, pages
  9334--9345. PMLR, 2020.

\bibitem{FernandezAl2016}
M~Giselle Fern{\'a}ndez-Godino, Chanyoung Park, Nam-Ho Kim, and Raphael~T
  Haftka.
\newblock Review of multi-fidelity models.
\newblock {\em arXiv preprint arXiv:1609.07196}, 2016.

\bibitem{ParkAl2017}
Chanyoung Park, Raphael~T Haftka, and Nam~H Kim.
\newblock Remarks on multi-fidelity surrogates.
\newblock {\em Structural and Multidisciplinary Optimization},
  55(3):1029--1050, 2017.

\bibitem{PoloczekAl2017}
Matthias Poloczek, Jialei Wang, and Peter~I Frazier.
\newblock Multi-information source optimization.
\newblock In {\em Proceedings of the 31st International Conference on Neural
  Information Processing Systems}, pages 4291--4301, 2017.

\bibitem{WuAl2020}
Jian Wu, Saul Toscano-Palmerin, Peter~I Frazier, and Andrew~Gordon Wilson.
\newblock Practical multi-fidelity bayesian optimization for hyperparameter
  tuning.
\newblock In {\em Uncertainty in Artificial Intelligence}, pages 788--798.
  PMLR, 2020.

\bibitem{LiAl2020}
Shibo Li, Wei Xing, Robert Kirby, and Shandian Zhe.
\newblock Multi-fidelity bayesian optimization via deep neural networks.
\newblock {\em Advances in Neural Information Processing Systems}, 33, 2020.

\bibitem{Ginsbourger&LeRiche2010}
David Ginsbourger and Rodolphe Le~Riche.
\newblock Towards gaussian process-based optimization with finite time horizon.
\newblock In {\em mODa 9--Advances in Model-Oriented Design and Analysis},
  pages 89--96. Springer, 2010.

\bibitem{LamAl2016}
Remi Lam, Karen Willcox, and David~H Wolpert.
\newblock Bayesian optimization with a finite budget: An approximate dynamic
  programming approach.
\newblock {\em Advances in Neural Information Processing Systems}, 29:883--891,
  2016.

\bibitem{Lam&Willcox2017}
Remi Lam and Karen Willcox.
\newblock Lookahead bayesian optimization with inequality constraints.
\newblock In {\em NIPS}, pages 1890--1900, 2017.

\bibitem{Powell2007}
Warren~B Powell.
\newblock {\em Approximate Dynamic Programming: Solving the curses of
  dimensionality}, volume 703.
\newblock John Wiley \& Sons, 2007.

\bibitem{OsborneAl2009}
Michael~A Osborne, Roman Garnett, and Stephen~J Roberts.
\newblock Gaussian processes for global optimization.
\newblock In {\em 3rd international conference on learning and intelligent
  optimization (LION3)}, pages 1--15. Citeseer, 2009.

\bibitem{GonzalezAl2016}
Javier Gonz{\'a}lez, Michael Osborne, and Neil Lawrence.
\newblock Glasses: Relieving the myopia of bayesian optimisation.
\newblock In {\em Artificial Intelligence and Statistics}, pages 790--799.
  PMLR, 2016.

\bibitem{Rasmussen2003}
Carl~Edward Rasmussen.
\newblock Gaussian processes in machine learning.
\newblock In {\em Summer school on machine learning}, pages 63--71. Springer,
  2003.

\bibitem{JonesAl1998}
Donald~R Jones, Matthias Schonlau, and William~J Welch.
\newblock Efficient global optimization of expensive black-box functions.
\newblock {\em Journal of Global optimization}, 13(4):455--492, 1998.

\bibitem{SrinivasAl2009}
Niranjan Srinivas, Andreas Krause, Sham~M Kakade, and Matthias Seeger.
\newblock Gaussian process optimization in the bandit setting: No regret and
  experimental design.
\newblock {\em arXiv preprint arXiv:0912.3995}, 2009.

\bibitem{Hennig&Schuler2012}
Philipp Hennig and Christian~J Schuler.
\newblock Entropy search for information-efficient global optimization.
\newblock {\em Journal of Machine Learning Research}, 13(6), 2012.

\bibitem{Wang&Jegelka2017}
Zi~Wang and Stefanie Jegelka.
\newblock Max-value entropy search for efficient bayesian optimization.
\newblock In {\em International Conference on Machine Learning}, pages
  3627--3635. PMLR, 2017.

\bibitem{HernandezAl2014}
Jos{\'e}~Miguel Hern{\'a}ndez-Lobato, Matthew~W Hoffman, and Zoubin Ghahramani.
\newblock Predictive entropy search for efficient global optimization of
  black-box functions.
\newblock {\em arXiv preprint arXiv:1406.2541}, 2014.

\bibitem{Bertsekas1995}
Dimitri Bertsekas.
\newblock {\em Dynamic programming and optimal control: Volume I}, volume~1.
\newblock Athena scientific, 1995.

\bibitem{WilsonAl2018}
James~T Wilson, Frank Hutter, and Marc~Peter Deisenroth.
\newblock Maximizing acquisition functions for bayesian optimization.
\newblock {\em arXiv preprint arXiv:1805.10196}, 2018.

\bibitem{Mckay1992}
Michael~D McKay.
\newblock Latin hypercube sampling as a tool in uncertainty analysis of
  computer models.
\newblock In {\em Proceedings of the 24th conference on Winter simulation},
  pages 557--564, 1992.

\bibitem{ForresterAl2008}
Alexander Forrester, Andras Sobester, and Andy Keane.
\newblock {\em Engineering design via surrogate modelling: a practical guide}.
\newblock John Wiley \& Sons, 2008.

\bibitem{MaininiAl2022}
L~Mainini, A~Serani, MP~Rumpfkeil, E~Minisci, D~Quagliarella, H~Pehlivan,
  S~Yildiz, S~Ficini, R~Pellegrini, F~Di~Fiore, et~al.
\newblock Analytical benchmark problems for multifidelity optimization methods.
\newblock {\em arXiv preprint arXiv:2204.07867}, 2022.

\bibitem{XiongAl2013}
Shifeng Xiong, Peter~ZG Qian, and CF~Jeff Wu.
\newblock Sequential design and analysis of high-accuracy and low-accuracy
  computer codes.
\newblock {\em Technometrics}, 55(1):37--46, 2013.

\end{thebibliography}

\end{document}